# Evaluating the Impact of AI-Powered Audiovisual Personalization on Learner Emotion, Focus, and Learning Outcomes

GEORGE WANG* and JINGYING DENG*, New York University, USA
SAFINAH ALI, New York Unveristy, USA

Independent learners often struggle with sustaining focus and emotional regulation in unstructured or distracting settings. Although some rely on ambient aids such as music, ASMR, or visual backgrounds to support concentration, these tools are rarely integrated into cohesive, learner-centered systems. Moreover, existing educational technologies focus primarily on content adaptation and feedback, overlooking the emotional and sensory context in which learning takes place. Large language models have demonstrated powerful multimodal capabilities including the ability to generate and adapt text, audio, and visual content. Educational research has yet to fully explore their potential in creating personalized audiovisual learning environments. To address this gap, we introduce an AI-powered system that uses LLMs to generate personalized multisensory study environments. Users select or generate customized visual themes (e.g., abstract vs. realistic, static vs. animated) and auditory elements (e.g., white noise, ambient ASMR, familiar vs. novel sounds) to create immersive settings aimed at reducing distraction and enhancing emotional stability. Our primary research question investigates how combinations of personalized audiovisual elements affect learners' cognitive load and engagement. Using a mixed-methods design that incorporates biometric measures and performance outcomes, this study evaluates the effectiveness of LLM-driven sensory personalization. The findings aim to advance emotionally responsive educational technologies and extend the application of multimodal LLMs into the sensory dimension of self-directed learning.

## 1 Introduction

### 1.1 Target Audience

The Whisper project is designed for self-directed learners and professionals who often face challenges with focus and emotional regulation in distracting or unstructured environments[**article**]. For example, higher education students may require an enhanced focus environment for studying or academic tasks[Yuvaraj et al. 2025]. It also targets professionals seeking a virtual space to increase productivity and maintain focus, such as freelance graphic designers or accountants experiencing stress during high-stress work periods. Whisper aims to support underserved users, including those with limited access to quiet or safe learning spaces.

### 1.2 Identified Need

The core need identified is the challenge independent learners face with focus, attention, and emotional regulation in suboptimal study environments. Many learners attempt to improve their study spaces by searching online for aids like ASMR videos, white noise playlists, or relaxing images, a process that can be time-consuming and often yields unsatisfactory results[Othman et al. 2019]. There is a need for integrated tools that directly support emotional regulation and focus through personalized sensory experiences. This need is particularly pronounced for learners who are neurodivergent, easily distracted, or lack access to ideal study spaces[Chen et al. 2022].

*Both authors contributed equally to this research.

Authors' Contact Information: George Wang, xw3617@nyu.edu; Jingying Deng, jd5403@nyu.edu, New York University, New York, New York, USA; Safinah Ali, New York Unveristy, New York, New York, USA, sa9140@nyu.edu.

Furthermore, professionals also seek tailored environments to enhance productivity and manage stress[Deng et al. 2024].

### 1.3 Research Question

The primary research question guiding the project is: How do different combinations of personalized visual and auditory elements affect learners' cognitive load and engagement? The project aims to address the problem of independent learners facing challenges with attention and emotional regulation by introducing an AI-powered system for creating personalized multisensory learning environments.

### 1.4 Prototype Design

The Whisper system is an AI-powered study companion that harnesses the multi-model ability of Large Language Models to transform a user's environment into a personalized sanctuary for focus, emotion, and cognitive growth. It enables users to build personalized multisensory environments by pairing self-selected visual themes with ASMR-based audio elements. Users can input text to generate preferred images, which can be set as their desktop wallpaper. They can also use text or images to generate background music or white noise, specify the duration, and choose the type of audio. The system uses AI to generate music and images based on user preferences to create a virtual personalized learning space. LLM embedded assistance is crucial for tailoring these elements to individual preferences efficiently and adaptively. The system is designed to be lightweight and accessible to minimize digital inequality and support underserved users. Whisper operates on a user-initiated, consent-based model, ensuring human autonomy and transparency in data handling.

### 1.5 Findings

The efficacy of the AI-generated personalized visual and auditory environments is planned to be measured using a mixed-methods approach. The research aims to evaluate improvements in learning efficiency and emotional states, including focus, relaxation, and satisfaction. We plan tp use data collection methods such as biometric measures to analyze focus and engagement levels. Behavioral measures like performance-based quizzes are used to assess learning outcomes, specifically memory retention. Additionally, self-reported data is collected through physiological surveys (like BFI, administered pre- and post-test) to assess changes in emotional states and perceived stress.

### 1.6 Rationale for AI Assistance

AI assistance is critical in this context for several reasons. First, AI enables personalization, allowing both visual and auditory elements to be tailored to the user's specific preferences. This ensures a more engaging and effective learning experience, as users can interact with an environment that resonates with their individual needs and preferences[Kanchon et al. 2024]. Second, AI enhances efficiency by quickly generating personalized visuals and audio, saving users time and providing them with a flexible learning space that can adapt to their needs without requiring manual intervention[Maity and Deroy 2024]. Lastly, AI supports adaptive learning, as the system can continuously evolve based on user interactions and feedback, improving the personalization of the auditory-visual environment and the user's learning experience[Ainary 2025].

# 2 System Design

## 2.1 Feature Mapping Based on User Needs

Based on the identified user needs, we developed a feature mapping table (Table 1) to systematically translate user challenges into corresponding design goals and system functionalities.

Table 1. Feature Mapping Based on User Needs

| User Needs | Design Goals | Mapped Features |
|---|---|---|
| Some independent learners struggle to maintain focus during study sessions. | Improve focus and attention | *Timer* and *Learning Environment* modules |
| Learners spend excessive time searching for external aids like ASMR videos. | Reduce time spent on setup | AI-generated personalized environment |
| Need for tools that support emotional regulation via multimodal stimuli. | Integrate multimodal regulation | *Whisper System* combining visual, auditory, and interactive elements |

- **Visual:** AI-generated personalized images are created based on user input and set as the desktop wallpaper. Users can select from various styles, including realistic, abstract, and artistic themes. These visuals are intended to enhance emotional engagement and maintain attention during study sessions.
- **Auditory:** AI-generated background music (BGM) and white noise are matched with the selected visual style. Users can choose between different types of audio, including white noise, ambient music, and layered sounds, to construct a customized auditory environment that supports focus and emotional regulation.
- **Interactive:** Users interact with the system via a text input interface to generate both visual and auditory elements. They can also control parameters such as music type, duration, and intensity, ensuring that both sensory modalities align with personal preferences and learning needs.

## 2.2 Iterative Prototyping Process

The development of the Whisper system followed an iterative prototyping methodology:

*2.2.1 First Prototype (Canvas Wireframe).* An initial static wireframe was created using Canvas. This prototype focused purely on layout planning and did not include any interactive features.

*2.2.2 Second Prototype (WIX Implementation).* To simulate interactivity, the second prototype was developed on WIX. While still using pre-assigned visual and auditory outputs, it introduced basic user controls, mimicking the intended experience.

*2.2.3 Third Prototype (Functionality Integration).* George implemented a functional backend while Crystal translated the WIX layout into a coded frontend. This version included:

- Text-to-image generation
- A draw mode (non-generative)
- Audio generation based on text prompts
- AI use agreement blocker: Users must accept terms before accessing key functions

Limitations included poor UI positioning and incomplete integration of drawing functionality.

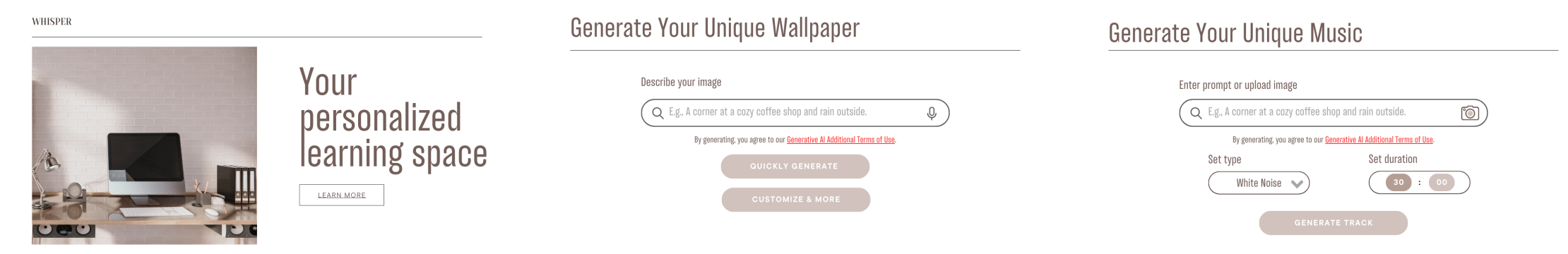

(a) Prototype 1: Canvas Wireframes (b) Prototype 1: Wallpaper (c) Prototype 1: Music

Fig. 1. Overview of Prototype 1 with Canvas wireframes, image preview, and music setup.

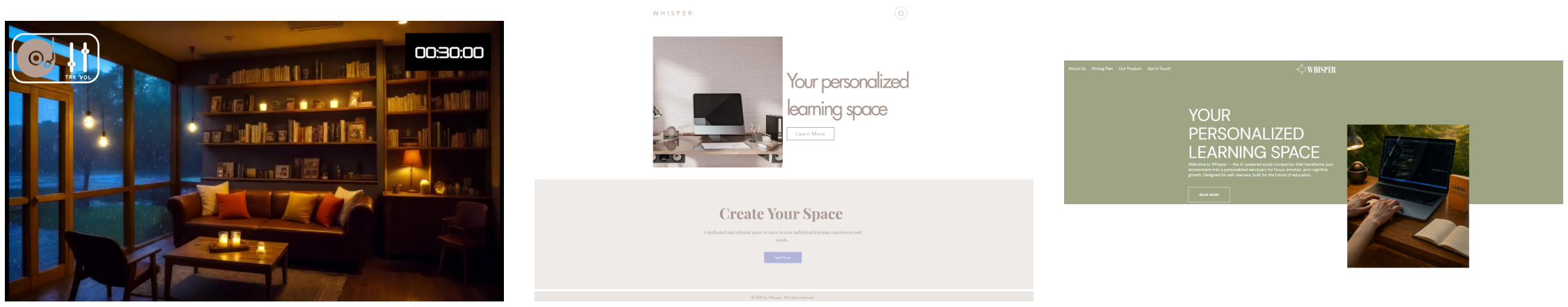

(a) Prototype 2: Wallpaper (b) Prototype 2: Starter (c) Prototype 2: Music

Fig. 2. Overview of Prototype 2 with WIX

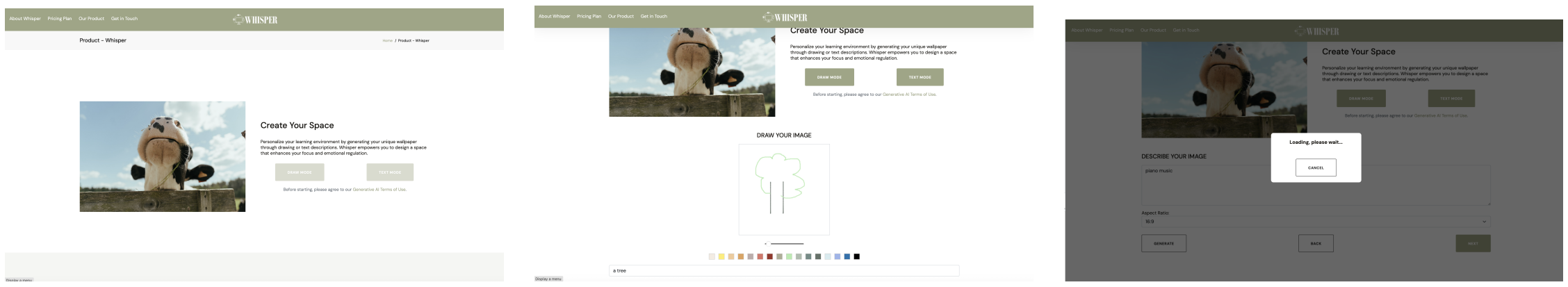

(a) Prototype 3: Overview (b) Prototype 3: Wallpaper (c) Prototype 3: Music Generation

Fig. 3. Overview of Prototype 3 with Functionality Integration

*2.2.4 Fourth Prototype (Full Integration).* The final prototype incorporated all initially planned features. Crystal redesigned the layout to improve usability, accessibility, and aesthetic polish.

### 2.3 Design Refinements and Feedback Integration

- **Draw Mode:** Added upon Prof. Safinah Ali's recommendation to support abstract, nonverbal thinking styles.
- **"None" Music Option:** Introduced based on user feedback (Zulsyika, Danira) to accommodate users preferring silent study environments.
- **Layout Redesign:** Following George's analysis of peer projects, the interface was refined to resolve visual and navigation issues.

## 3 Prototypes

The final prototype of the Whisper system is a fully responsive, browser-based interactive environment designed to enhance focus and emotional regulation through personalized AI-generated visual and auditory stimuli. It integrates multiple AI-driven features, including text-to-image, image-to-image, text-to-music, and image-to-music functionalities, with a structured user interface and accessibility controls.

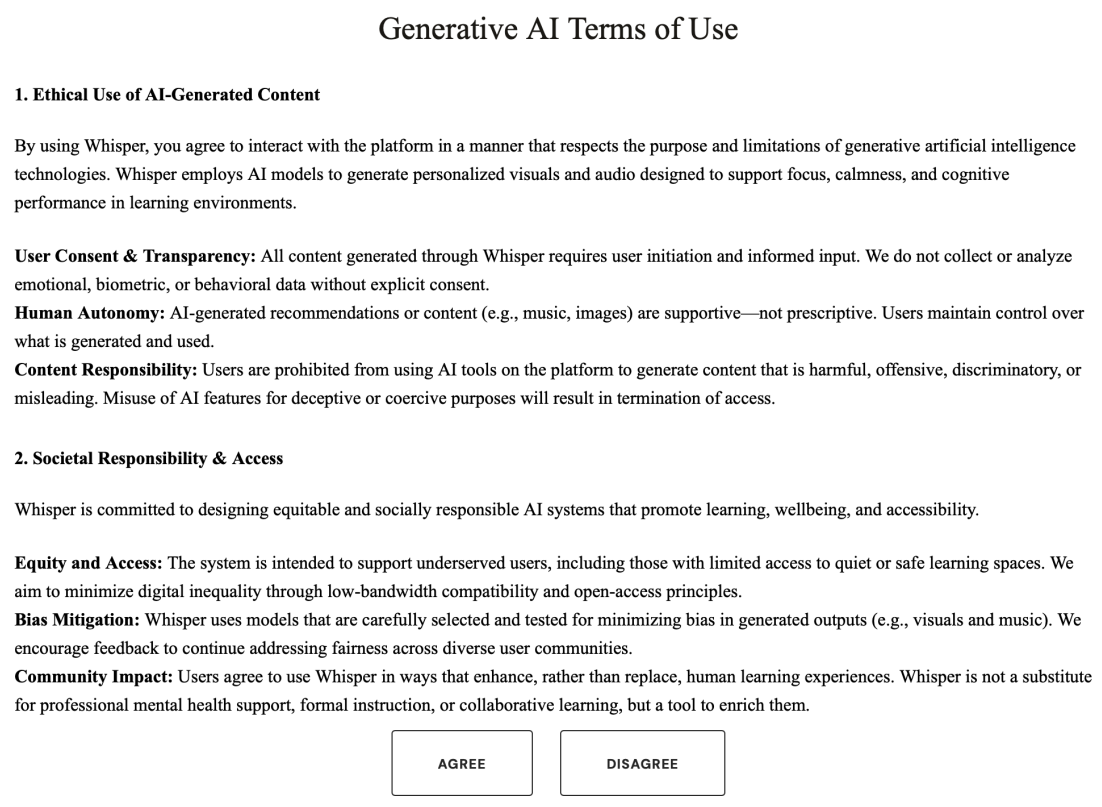

Fig. 4. AI Policy Enforcement

## 3.1 AI Policy Enforcement

To ensure ethical engagement with AI-generated content, a blocking mechanism is implemented. Users must first agree to the Terms of AI Use before gaining access to any AI-driven functions. Until this agreement is accepted, all AI generation features—including both visual and auditory content—remain disabled.

## 3.2 Image Generator

The image generator provides two distinct modes to accommodate different user expression styles:

*3.2.1 Text-to-Image Mode (Text Mode).* Users input a descriptive prompt (e.g., "a peaceful mountain cabin at sunset") into a text field. This input is processed by the ChatGPT image generation backend, which returns a high-quality image reflecting the requested scene. This mode is optimized for users who prefer verbal expression or quick ideation without manual sketching.

*3.2.2 Image-to-Image Mode (Draw Mode).* This mode provides users with a canvas for drawing their ideas directly. It includes:

- **Brush Tool:** Adjustable brush size for expressive flexibility.
- **Eraser Tool:** Enables precise removal of unwanted elements.
- **Color Palette:** Includes 16 curated colors selected for emotional and aesthetic harmony.
- **Brush Size Slider:** Offers fine-grained control over stroke thickness.

After drawing, the sketch is processed by the Gemini model, which interprets the visual input and generates a refined, high-resolution image. This mode is ideal for users who prefer visual or tactile interaction.

## 3.3 Music Generator

The music generation module is designed to create ambient audio experiences aligned with the emotional or thematic tone of the user's generated visuals or textual prompts.

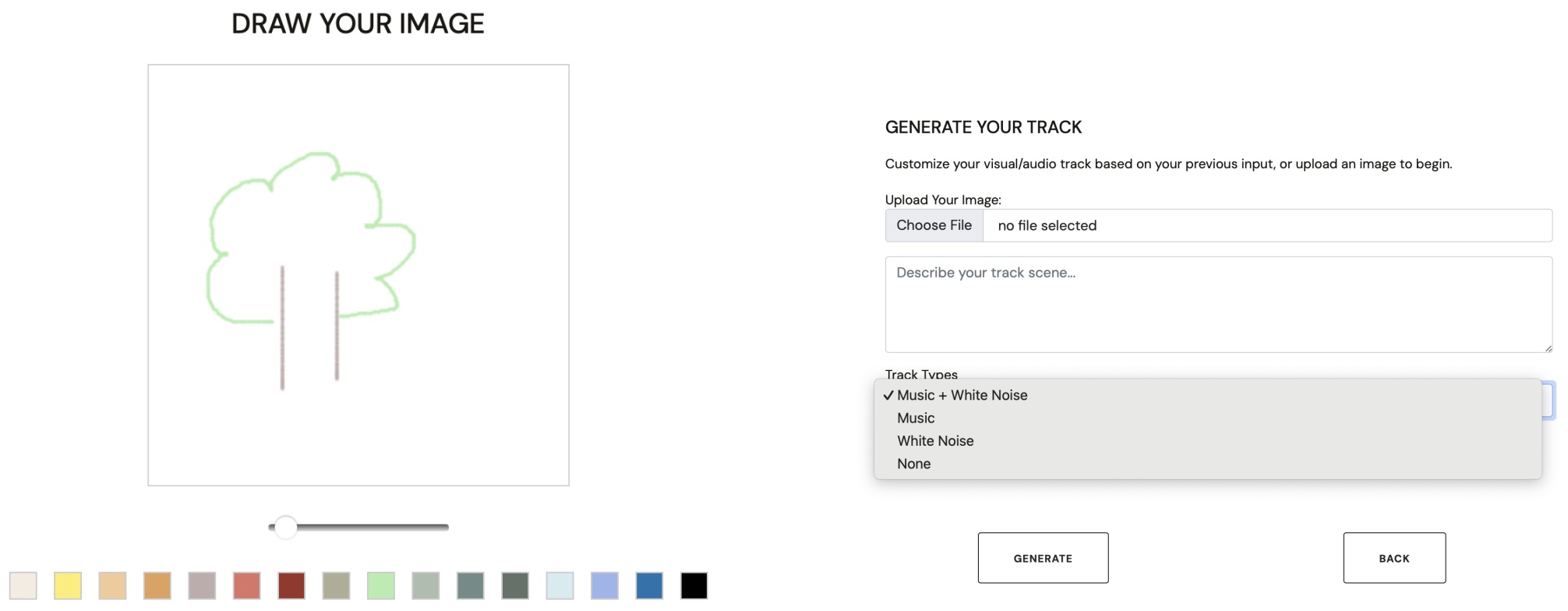

(a) Draw-to-image interface. (b) Prompt-based image generator.

Fig. 5. Visual customization modes in the Whisper prototype.

*3.3.1 Image-to-Music Mode.* Users select a previously generated image and may optionally add a mood descriptor (e.g., "serene," "energetic"). The system analyzes the visual content and generates music using the MusicGen model, mapping visual aesthetics to auditory parameters.

*3.3.2 Text-to-Music Mode.* Users enter prompts such as "lo-fi jazz" or "forest sounds with piano." The MusicGen model interprets the input using natural language understanding to generate audio that fits the desired mood or genre.

*3.3.3 Music Settings.*

- **Audio Type Selector:** Offers four modes—Music + White Noise, Music only, White Noise only, or None (for users who prefer silence).
- **Duration Selector:** Allows users to set playback length from 1 second to 100 minutes.

Generated music is automatically loaded into the final environment and synchronized with the visual background.

### 3.4 Learning Environment

The learning environment provides an immersive, multisensory space where users engage in focused work sessions:

- **Visual Background:** The selected AI-generated image fills the screen to set the atmosphere.
- **Music Player:** Located at the bottom of the interface, featuring:
  - Play/Pause toggle
  - Volume slider
  - Speed adjustment (0.5x to 3x)
  - Download option
  - Automatic looping synced with timer duration
- **Timer:** A customizable countdown timer (e.g., 25 minutes) is displayed at the top-right. When the timer ends, the music automatically stops.
- **Navigation Controls:** A "Back" button in the top-left corner returns users to the generation interface for iteration.

## 4 Evaluation

The evaluation of the Whisper system was designed to rigorously assess its effectiveness in supporting cognitive focus, emotional regulation, and user experience during self-directed learning tasks. This system uses AI-generated personalized visual and auditory environments to enhance learner engagement. The evaluation follows a mixed-methods approach combining biometric data, behavioral observation, and qualitative feedback to form a multi-dimensional understanding of system impact.

### 4.1 Evaluation Environment and Procedure

The study was conducted in a controlled research room equipped with specialized data collection tools. Fig. 6 presents the layout of the room. The participant interacts with the system at a desk outfitted with a computer, a front-facing camera for facial expression recognition (FER), and an eye-tracking sensor. A smartphone positioned nearby is used for audio recording during post-task interviews. A researcher is present in the room to oversee the session without interfering with the participant's interaction[Geraets et al. 2021].

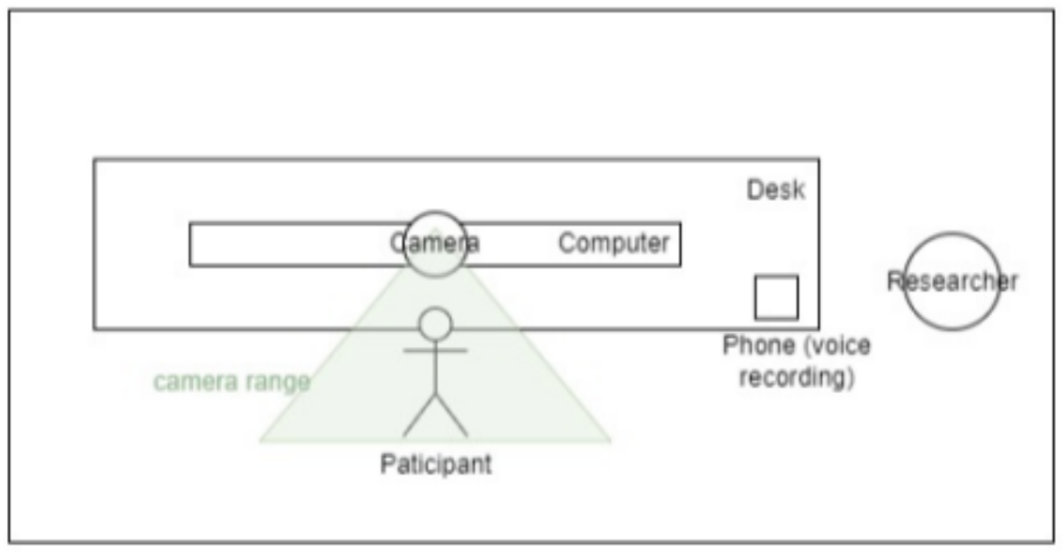

Fig. 6. Research room layout used during evaluation.

Each evaluation session follows this sequence:

(1) **Pre-Test Survey:** Participants complete the Big Five Inventory (BFI) and a short cognitive load and stress-level self-assessment.
(2) **System Interaction:** Participants use the Whisper system to create a personalized environment and engage in a 20–30 minute reading and memorization task.
(3) **Post-Test Activities:** Participants complete a quiz testing information retention, followed by self-report emotional assessments, and a semi-structured interview[Kosel et al. 2024].

### 4.2 Evaluation Metrics

To comprehensively understand the system's impact, we focus on three categories: cognitive focus, emotional state, and experiential satisfaction. These categories are measured using a combination of sensors, self-reports, and qualitative interviews, as detailed in Table 2.

### 4.3 Data Collection Methods

The evaluation utilizes a multi-layered data collection approach:

*4.3.1 Behavioral and Biometric Observation.* Eye-tracking data is collected to determine participant focus and engagement by monitoring gaze fixation, saccades, and dwell time on key screen elements. Facial expression recognition (FER) captures changes in emotional state throughout the task. Participant body posture and hand gestures are recorded and coded for indicators of anxiety, engagement, or fatigue.

Table 2. Evaluation Metrics and Measurement Methods

| Category | What We Measure | How We Measure It |
|---|---|---|
| Cognitive Focus | Time-on-task, attention patterns | Eye-tracking data (dwell time, fixations), self-rated focus score |
| Emotional State | Calmness, relaxation, stress | Facial expression analysis (FER), emotion sliders, posture/body language |
| Experience Value | Satisfaction with visual/audio design | Post-task interviews, thematic analysis, "vibe check" questions |

*4.3.2 Surveys and Self-Reports.* Participants complete pre-test and post-test surveys to reflect on their emotional state and perceived learning efficiency. These include the Big Five Inventory (BFI), self-rated focus scales, and a post-session quiz testing vocabulary and reading retention. Additional Likert-scale questions assess usability and comfort with the system.

*4.3.3 Qualitative Interviews.* Each participant undergoes a semi-structured interview immediately after system use. Interview topics include ease of use, emotional impact of the generated environment, perceived productivity, and preferences regarding image and music customization. Responses are audio recorded for transcription and thematic coding[Skaramagkas et al. 2023].

### 4.4 Triangulation of Data

To enhance reliability, results from different methods are triangulated. For instance, if eye-tracking data shows prolonged fixation and self-reports confirm focus, the measurement is validated. Discrepancies are explored through follow-up questioning or reanalysis of FER/emotion slider logs[Li et al. 2025].

### 4.5 Limitations

Although this evaluation framework is comprehensive, it is currently theoretical. Pilot results and real user data are not yet available. Future work will involve user testing with diverse populations, including neurodivergent learners, to validate findings across contexts.

## 5 Ethical Implications

### 5.1 User Privacy and Consent

Whisper is built with a strong focus on ethical responsibility, especially when using AI in learning environments. A significant concern is user privacy, particularly regarding the collection of personal information such as emotional states, preferences, or behavior. To protect users, Whisper employs a transparent and consent-based approach. Before using the system, users are introduced to our "Generative AI Terms of Use," which explains how the AI functions and how data is managed. No data is collected without explicit permission. If users choose, their inputs can be stored locally to mitigate privacy risks. Whisper's AI tools are always user-initiated, ensuring that the system does not make automatic decisions or changes without the user's consent. This design ensures that learners maintain control over their environment.

This approach aligns with emerging frameworks emphasizing user control and transparency in AI systems. For instance, the Privacy Ethics Alignment in AI (PEA-AI) model advocates for stakeholder-centric privacy governance, highlighting the importance of informed consent and user autonomy in AI-driven environments [Liyanarachchi et al. 2025].

### 5.2 Emotional Autonomy and Avoidance of Manipulation

Whisper avoids emotional manipulation by refraining from hidden mood detection or behavior tracking. Instead, it offers tools to support emotional regulation. By granting users control over when and how they utilize AI features, Whisper respects human agency. The goal is not to replace a learner's decisions but to assist them in making better choices with reduced stress. This approach helps mitigate potential harms from AI and fosters trust between the user and the system.

Research supports the importance of user autonomy in emotional regulation within AI systems. For example, the Reflexion platform integrates real-time emotion recognition with user-driven reflective prompts to enhance emotional literacy without compromising user agency [Wang et al. 2025].

### 5.3 Societal Implications and Digital Equity

Whisper supports learners who often struggle with focus and motivation, particularly those studying alone or in noisy, uncomfortable environments. Many students attempt to improve their study settings by searching for suitable ASMR videos, white noise playlists, or relaxing images online. Whisper alleviates this burden by using AI to generate personalized music and images based on user preferences, helping learners achieve a calm and focused state without the time-consuming search. This can be especially beneficial for neurodivergent learners, those easily distracted, or individuals with limited access to ideal study spaces.

Furthermore, Whisper contributes to reducing digital inequality. It is designed to be lightweight and accessible, even for users with limited resources. By providing personalized study support at no cost, Whisper fosters a more inclusive and supportive learning environment. This demonstrates how AI, when used responsibly, can enhance educational equity and effectiveness for all learners.

Studies have highlighted the role of AI in promoting inclusive education. For instance, AI-driven tools have been shown to support learners with special needs and tackle digital inequality by providing tailored educational resources [Fan et al. 2024].

### Acknowledgments

The authors would like to express their sincere gratitude to Prof. Safinah Ali for her insightful guidance and encouragement throughout the development of this project. We also thank the Experience Design and Artificial Intelligence (XDAI) course community for fostering a supportive and inspiring learning environment. This work would not have been possible without the collaboration, feedback, and shared creativity of everyone in the class.